\documentclass[11pt,a4paper]{article}
\usepackage[hyperref]{eacl2021}
\usepackage{times}
\usepackage{latexsym}

\usepackage{microtype}

\aclfinalcopy 

\usepackage{booktabs}
\usepackage{makecell}
\usepackage{tabularx}
\usepackage{multirow}
\usepackage{arydshln}
\usepackage{tabu}
\usepackage{xcolor}

\usepackage{amsmath}
\usepackage{amssymb}
\usepackage{bm}

\usepackage{ dsfont }

\usepackage{microtype}

\usepackage{enumitem}

\usepackage{graphicx}
\usepackage{caption}

\usepackage{esvect}

\usepackage{anyfontsize}
-lmtk10

\usepackage{color}

\makeatletter
\newcommand{\smallsym}[2]{#1{\mathpalette\make@small@sym{#2}}}
\newcommand{\make@small@sym}[2]{%
  \vcenter{\hbox{$\m@th\downgrade@style#1#2$}}%
}
\newcommand{\downgrade@style}[1]{%
  \ifx#1\displaystyle\scriptstyle\else
    \ifx#1\textstyle\scriptstyle\else
      \scriptscriptstyle
  \fi\fi
}
\makeatother

\makeatletter
  \newcommand\smallernormal{\@setfontsize\smallernormal{9.5pt}{9.5}}
\makeatother
\makeatletter
  \newcommand\smallernormalAppendix{\@setfontsize\smallernormalAppendix{11pt}{10}}
\makeatother

\newcommand{\yulan}[1]{\textcolor{black}{{#1}}}
\newcommand{\ek}[1]{\textcolor{black}{{#1}}}
\newcommand{\ml}[1]{\textcolor{black}{{#1}}}
\title{Boosting Low-Resource Biomedical QA \\ via Entity-Aware Masking Strategies}

\author{
Gabriele Pergola$^1$, Elena Kochkina$^{1,3}$, Lin Gui$^1$, Maria Liakata$^{1,2,3}$, Yulan He$^1$\\
$^1$University of Warwick, UK $\quad$
$^2$Queen Mary University of London, UK\\
$^3$The Alan Turing Institute, UK\\
\texttt{\{gabriele.pergola,e.kochkina,lin.gui,yulan.he\}@warwick.ac.uk}\\
\texttt{m.liakata@qmul.ac.uk}
}

\date{}

\begin{document}
\maketitle

\begin{abstract}
Biomedical question-answering (QA) has gained increased attention for its capability to provide users with high-quality information from a vast scientific literature. Although an increasing number of biomedical QA datasets has been recently made available, those resources are still rather limited and expensive to produce. 
Transfer learning via pre-trained language models (LMs) has been shown as a promising approach to leverage existing general-purpose knowledge. 
However, fine-tuning these large models can be costly and time consuming\ml{,} 
often \ml{yielding} limited benefits when adapting to specific themes of specialised domains, such as the COVID-19 literature.
\ml{To} bootstrap further their domain adaptation, we propose a simple yet unexplored approach, which we call \textit{biomedical entity-aware masking} (BEM)\ml{.} 
\ml{We encourage} masked language models to learn entity-centric knowledge based on the pivotal entities characterizing the domain at hand, and employ those entities to drive the LM fine-tuning.
The resulting strategy is a downstream process applicable to a wide variety of masked LMs, not requiring additional memory or components in the neural architectures. Experimental results show performance on par with 
state-of-the-art models on several biomedical QA datasets. 
\end{abstract}

\section{Introduction}

Biomedical question-answering (QA) \yulan{aims to provide users with succinct answers given their queries by analysing a large-scale scientific literature.} 
\yulan{It enables} clinicians, public health officials and end-users 
to 
\yulan{quickly} access the rapid flow of specialised knowledge continuously produced. 
This has led the research community's effort towards \yulan{developing} specialised models and tools \yulan{for biomedical QA and} 
assessing their performance \yulan{on benchmark datasets such as  BioASQ} \cite{Tsatsaronis15}. 
%
Producing such data is time-consuming and requires involving domain experts, making it an expensive process. As a result, high-quality biomedical QA datasets are a scarce resource.
\ml{The} recently released CovidQA collection \cite{tang20}, the first manually curated dataset about COVID-19 related issues, provides only 127 question-answer pairs.
Even one of the largest available biomedical QA datasets, BioASQ, only contains a few thousand 
questions.  

\begin{figure}[!t]
\centering
\includegraphics[width=1.0\columnwidth]{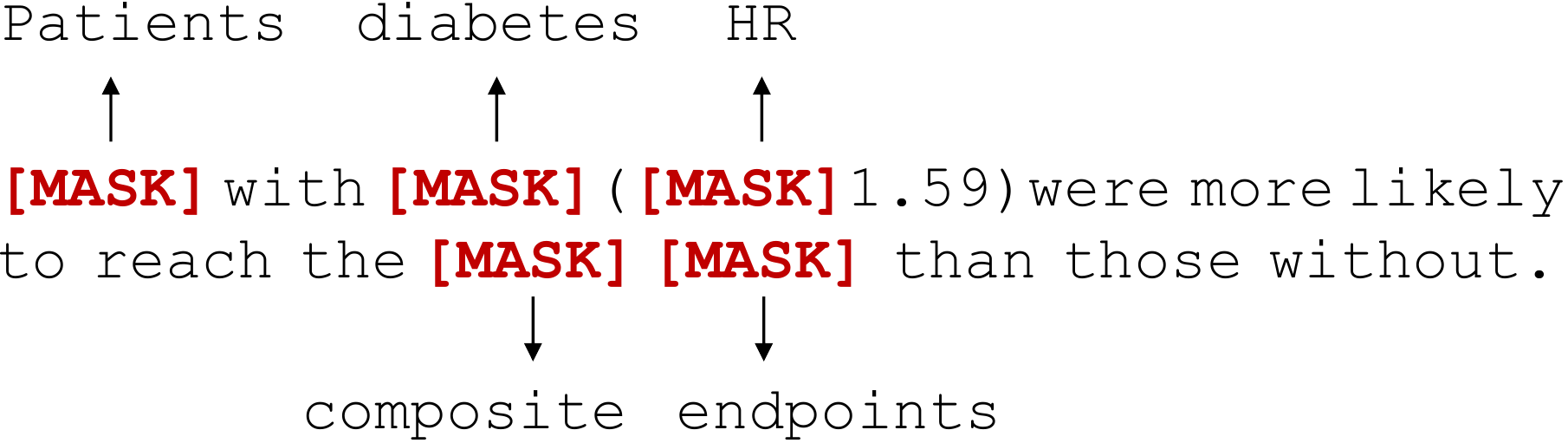}
\caption{An excerpt of a sentence masked via the BEM strategy, where the masked words were chosen through a biomedical named entity recognizer. In contrast, BERT \cite{devlin19} would randomly select the words to be masked, without attention to the relevant concepts characterizing a technical domain.} 
\label{fig:bem_sent_masked}
\end{figure}

\yulan{There have been attempts to fine-tune pre-trained large-scale language models} 
for general-purpose QA tasks \cite{rajpurkar16, liu19, raffel20} \yulan{and then use them directly for biomedical QA}. 
\yulan{Furthermore, there \ml{has also been} increasing interest in developing domain-specific} 
language models, such as BioBERT \cite{Lee19} or RoBERTa-Biomed \cite{gururangan20}, leveraging the vast medical literature available.
\ek{While achieving state-of-the-art results on \ml{the} QA task, these models come with} a high computational cost: BioBERT \ml{needs} ten days on eight GPUs to train \cite{Lee19}, making it prohibitive for 
researchers \yulan{with no access to massive computing resources}. 

\begin{figure*}[!t]
\centering
\includegraphics[width=0.96\textwidth]{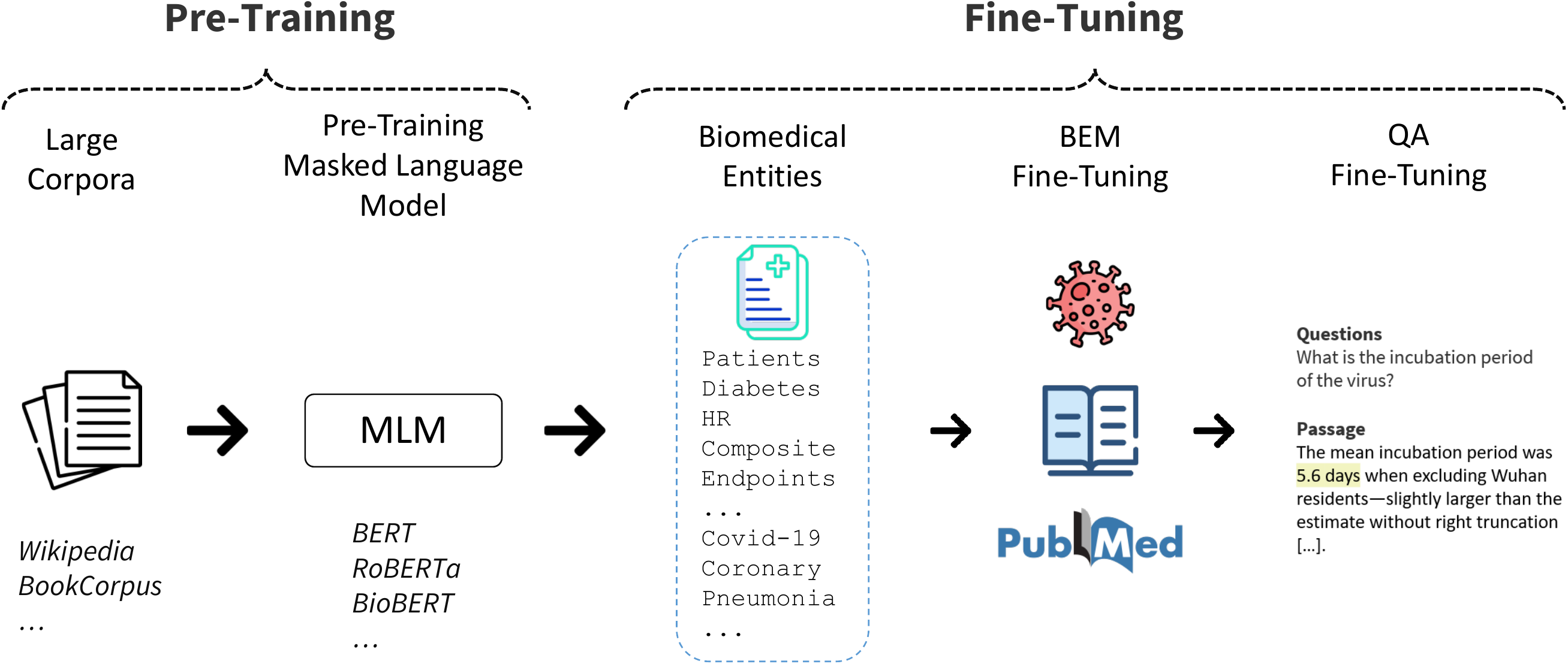}
\caption{A schematic representation of the main steps involved in fine-tuning masked language models for the QA task through the biomedical entity-aware masking (BEM) strategy.} 
\label{fig:bem_diagram}
\end{figure*}

An alternative \ek{approach} to incorporating external knowledge \ek{into pre-trained language models} is to drive the LM \ek{to \ml{focus} on} pivotal entities characterising the domain at hand \ek{during the fine-tuning stage}. Similar ideas were explored in works by~\citet{zhang19}, \citet{Sun20}, which proposed the ERNIE model. However, their adaptation strategy was designed to generally improve the LM representations rather than adapting it to a \yulan{particular} 
domain, requiring additional objective functions and memory.
%
\ek{In this work we aim to enrich } existing general-purpose LM models (e.g. BERT\ml{~\cite{devlin19}}) with the knowledge related to key medical concepts. 
\yulan{In addition}, we want \yulan{domain-specific} 
LMs (e.g. BioBERT) to re-encode the already acquired information around the medical entities of interests for a particular \yulan{topic or} theme 
(e.g. literature \yulan{relating to COVID-19}).

Therefore, to \ek{facilitate} further domain adaptation, we propose a simple yet unexplored approach based on a novel masking strategy to fine-tune a LM. Our approach introduces a \textit{biomedical entity-aware masking} (BEM) strategy encouraging masked language models (MLMs) to learn entity-centric knowledge (\textsection \ref{subsect:bem}). We first identify a set of entities characterising the domain at hand using a \yulan{domain-specific} 
entity recogniser (SciSpacy \cite{neumann19}), and then employ a subset of those entities to drive the masking strategy while fine-tuning (Figure \ref{fig:bem_sent_masked}).
The resulting BEM strategy is 
applicable to a vast variety of MLMs and does not require additional memory or components in the neural architectures. Experimental results show performance on a par with the state-of-the-art models for biomedical QA tasks 
(\textsection \ref{sec:exp_results}) on several biomedical QA datasets. A further qualitative assessment provides an insight into how QA pairs benefit from the proposed approach. 



\begin{table*}[!htb]
\centering
\scalebox{0.9}{
\begin{tabular*}{0.83\textwidth}{@{\extracolsep{\fill}}*{8}{>{\smallernormal}l}}
\toprule
\multirow{2}{*}{\textbf{\#}} & \multirow{2}{*}{\textbf{Model}}  & \multicolumn{3}{c}{\textbf{CovidQA}} & \multicolumn{3}{c}{\textbf{BioASQ 7b}} \\ \cline{3-5} \cline{6-8}  
                            &  & P@1 & R@3 & MRR           & SAcc & LAcc & MRR \\ \toprule
1 $\ \ $ & \textbf{BERT} 							& 0.081$^*$ & 0.117$^*$  & 0.159$^*$ & 0.012 & 0.032  & 0.027	\\
2 & $\ \ $ \texttt{+ BioASQ}                & 0.125		&  0.177  & 0.206  	& 0.226  	& 0.317   & 0.262  	\\
3 & $\ \ $ \texttt{+ STM + BioASQ}         	& 0.132		&  0.195  & 0.218  	& 0.233  	& 0.325   & 0.265  	\\
4 & $\ \ $ \texttt{+ BEM + BioASQ}          & 0.145     &  0.278  & 0.269  	& 0.241 	& 0.341   & 0.288  	\\ \hline

5 & \textbf{RoBERTa} 		   				& 0.068  	& 0.115	  & 0.122   & 0.023 	& 0.041  & 0.036   	\\
6 & $\ \ $ \texttt{+ BioASQ}                & 0.106		& 0.155   & 0.178	& 0.278  	& 0.324	 & 0.294	\\
7 & $\ \ $ \texttt{+ STM + BioASQ}          & 0.112		& 0.167   & 0.194	& 0.282  	& 0.333	 & 0.300	\\
  
8 & $\ \ $ \texttt{+ BEM + BioASQ}          & 0.125   	& 0.198   & 0.236 	& 0.323 	& 0.374  & 0.325   	\\ \hline

9 & \textbf{RoBERTa-Biomed}					& 0.104  	& 0.163 	& 0.192		& 0.028 	& 0.044  	& 0.037		\\
10 & $\ \ $ \texttt{+ BioASQ}                & 0.128  	& 0.355  	& 0.315	   	& 0.415 	& 0.398		& 0.376		\\
11 & $\ \ $ \texttt{+ STM + BioASQ}          & 0.136  	& 0.364  	& 0.321	   	& 0.423 	& 0.410		& 0.397		\\
12 & $\ \ $ \texttt{+ BEM + BioASQ}          & 0.143  	& 0.386  	& 0.347		& \textbf{0.435} 	& 0.443   	& 0.398  	\\ \hline

13 & \textbf{BioBERT}  		   				& 0.097$^*$ 	& 0.142$^*$ & 0.170$^*$ 	& 0.031 	& 0.046  & 0.039	\\
14 & $\ \ $ \texttt{+ BioASQ}                & 0.166 		& 0.419    	& 0.348	   		& 0.410$^\dag$  & 0.474$^\dag$  & 0.409$^\dag$	\\
15 & $\ \ $ \texttt{+ STM + BioASQ}          & 0.172     	& 0.432  	& 0.385	  		& 0.418 	&  0.482  & 0.416  \\
16 & $\ \ $ \texttt{+ BEM + BioASQ} $\ \ \ $ & \textit{0.179}&\textbf{0.458} & \textit{0.391}     & 0.421 	&  \textbf{0.497}  & \textbf{0.434}  \\ \hline

17 & \textbf{T5 LM}  		   				   	\\	
18 & $\ \ $ \texttt{+ MS-MARCO} 			    & \textbf{0.282}$^*$ & 0.404$^*$ & \textbf{0.415}$^*$ 	& ---   	& ---     & ---		\\

\bottomrule
\end{tabular*}}
\caption{Performance of language models on the CovidQA and BioASQ 7b1 dataset. Values referenced with {\small*} come from the \citet{tang20} work and with \dag $\,$ from \citet{Yoon20}.}
\vspace{-6pt}
\label{tb:qa_results}
\end{table*}

\section{BEM: A Biomedical Entity-Aware Masking Strategy}

The fundamental principle of a masked language model (MLM) is to generate word representation\ml{s} that can be used to predict the missing tokens of an input text. 
While this general principle is adopted in the vast majority of MLMs, the particular way in which the tokens to be masked are chosen can vary  considerably. We thus proceed analysing the random masking strategy adopted in BERT \cite{devlin19} which has 
inspired most of the existing approaches, and we then introduce the biomedical entity-aware masking strategy used to fine-tune MLMs in the biomedical domain. 

\noindent \paragraph{BERT Masking strategy.}
The masking strategy adopted in BERT 
randomly replaces a predefined proportion of words with a special \texttt{[MASK]} token and the model is required to predict them. 
In BERT, 15\% of tokens are chosen uniformly at random, 10\% of them are swapped into random tokens (thus, resulting in an overall 1.5\% of the tokens randomly swapped). This introduces a rather limited \ml{amount of } noise with the aim of making the predictions more robust to trivial associations between the masked tokens and the context.
While another 10\% of the selected tokens are kept without modifications, the remaining 80\% of them are replaced with the \texttt{[MASK]} token.

\paragraph{Biomedical Entity-Aware Masking Strategy}
\label{subsect:bem}
We describe an entity-aware masking strategy \yulan{which only masks biomedical entities detected by a domain-specific} 
named entity recogniser (SciSpacy\footnote{https://scispacy.apps.allenai.org/}). 
Compared to the \ek{random masking} strategy 
described \ek{above, which is used} to pre-train the masked language models, 
the introduced entity-aware masking strategy is adopted to boost the fine-tuning process for biomedical documents. In this phase, rather than randomly choosing the tokens to be masked, we inform the model \ek{of} the relevant tokens to pay attention to, and \yulan{encourage the model} 
to refine \ml{its} representations using the new surrounding context. 

\paragraph{Replacing strategy}
We decompose the BEM strategy into two steps: (1) \textit{recognition} and (2) \textit{sub-sampling and substitution}.
During the \textit{recognition phase}, a set of biomedical entities $\mathcal{E}$ is identified in advance over a training corpus.

Then, at \ml{the} \textit{sub-sampling and substitution} stage, we first sample a proportion $\rho$ of biomedical entities $\mathcal{E_s} \in \mathcal{E}$. The resulting entity subsets $\mathcal{E_s}$ is thus dynamically computed at batch time, in order to introduce a diverse and flexible spectrum of masked entities during training. For consistency, we use the same tokeniser for the documents $d_i$ in the batch and the entities $e_j \in \mathcal{E}$. Then, we substitute all the $k$ entity mentions $w_{e_j}^{k}$ in $d_i$ with the special token \texttt{[MASK]}, making sure that no consecutive entities are replaced. 
The substitution takes place at batch time, so that the substitution is a downstream process suitable for a wide typology of MLMs.
A diagram synthesizing the involved steps is reported in Figure \ref{fig:bem_diagram}.



\begin{table*}[htbp]
\centering
\setlength\extrarowheight{2pt} 
\resizebox{\linewidth}{!}{%
\begin{tabular}{cl} 
\hline\hline
\textbf{BERT with STM} & \multicolumn{1}{c}{\textbf{BERT with BEM}} \\ 
\hline\hline
\multicolumn{2}{c}{\textit{What is the \textbf{OR }for severe infection in COVID-19 patients with hypertension?} } \\ 
\cmidrule(lr){1-2}
\multicolumn{1}{l}{\begin{tabular}[c]{@{}l@{}}\textbf{-} There were significant correlations between COVID-19 severity~\\and~[..],~diabetes [OR=2.67],~coronary heart~disease~[OR=2.85].\end{tabular}} & \begin{tabular}[c]{@{}l@{}}\textbf{-} There were significant correlations between COVID-19 severity~\\and [..],~diabetes [OR=2.67],~coronary heart disease~[OR=2.85].\end{tabular} \\
\multicolumn{1}{l}{\begin{tabular}[c]{@{}l@{}}\textbf{-} Compared with the non-severe patient, the pooled odds ratio of\\hypertension, respiratory system disease, cardiovascular disease in\\severe patients were (OR 2.36, ..), (OR 2.46, ..) and (OR 3.42, ..).\end{tabular}} & \begin{tabular}[c]{@{}l@{}}\textbf{-} Compared with the non-severe patient, the pooled odds ratio of\\hypertension, respiratory system disease, cardiovascular disease in\\severe patients were (OR 2.36, ..), (OR 2.46, ..) and (OR 3.42, ..).\end{tabular} \\ 
\cmidrule(r){1-2}
\multicolumn{2}{c}{\textit{What is the \textbf{HR }for severe infection in COVID-19 patients with hypertension?}} \\ 
\cmidrule(r){1-2}
- - - - & \begin{tabular}[c]{@{}l@{}}\textbf{-} After adjusting for age and smoking status, patients with COPD~\\(HR 2.681), diabetes (HR 1.59), and malignancy (HR 3.50) were \\ more likely to reach to the composite endpoints than those without.\end{tabular} \\ 
\cmidrule(r){1-2}
\multicolumn{2}{c}{\textit{What is the \textbf{RR }for severe infection in COVID-19 patients with hypertension?}} \\ 
\cmidrule(lr){1-2}
- - - - & \begin{tabular}[c]{@{}l@{}}\textbf{-} In univariate analyses, factors significantly associated with severe~\\COVID-19 were male sex (14 studies; pooled RR=1.70, ...),~hyper-\\tension (10 studies 2.74 ...),diabetes~(11 studies ...), and CVD (..).\end{tabular} \\
\bottomrule
\end{tabular}
}
\caption{Examples of questions and retrieved answers using BERT fine-tuned either with its original masking approach or with the biomedical entity-aware masking (BEM) strategy.}
\vspace{-7pt}
\label{tab:qa_pairs_codvidqa}
\end{table*}

\section{Evaluation Design}

\noindent\textbf{Biomedical Reading Comprehension}.
We represent a document as $d_i := (s_{0}^i,\ .\ .\ ,s_{j-1}^i)$ 
\ml{,} a sequence of sentences, \ml{in turn} defined as $s_j := (w_{0}^j,\ .\ .\ ,w_{k-1}^j)$, with $w_k$ a word occurring in $s_j$.
Given a question $q$, the task is to retrieve the span $w_{s}^j,\ .\ .\ ,w_{s+t}^j$ from a document $d_j$ that can answer the question.
We assume the extractive QA setting where the answer span to be extracted lies entirely within one, or more than one document $d_i$.

In addition, for consistency with \ml{the} CovidQA dataset and \ml{to compare with} 
results in \citet{tang20}, we consider a further and sightly modified setting in which the task consists of retrieving the sentence $s_{j}^i$ that most likely contain\ml{s} the exact answer. This sentence level QA task mitigates the non-trivial ambiguities intrinsic to the definition of the exact span for an answer, an issue particularly relevant in the medical domain and well-know in the literature \cite{Voorhees99}\footnote{Consider, for instance, the following QA pair: \textit{``What is the incubation period of the virus?"}, \textit{``6.4 days (95\% 175 CI 5.3 to 7.6)"}, where a model returning just \textit{``6.4 days"} would be considered wrong.}.

\noindent\textbf{Datasets}. We assess the performance of the proposed masking strategies on two biomedical datasets: CovidQA and BioASQ.

\noindent\textbf{CovidQA} \cite{tang20} is a manually curated dataset  based on the AI2's COVID-19 Open Research Dataset \cite{Wang20}. \ml{It} consists of 127 question-answer pairs with 27 questions and 85 unique \ml{related articles}.
~\ek{This dataset is} too small for supervised training, but is a valuable resource for zero-shot evaluation to assess the unsupervised and transfer capability of models. 

\noindent\textbf{BioASQ}~\cite{Tsatsaronis15} is one of the larger biomedical QA datasets available with over 2000 question-answer pairs. To use it within the extractive questions answering framework, we convert the questions into the SQuAD dataset format~\cite{rajpurkar16}, consisting of question-answer pairs and the corresponding \textit{passages,} 
medical articles containing the answers or clues with a length varying from a sentence to a paragraph. 
When multiple passages are available for a single question, we form additional question-context pairs combined subsequently in a postprocessing step to choose the answer with highest probability, similarly to \citet{Yoon20}.
For consistency with the CovidQA dataset, we report our evaluation exclusively on the factoid questions of the BioASQ 7b Phase B1.

\noindent\textbf{Baselines}. \ek{We use the following unsupervised neural models as baselines:} the out-of-the-box BERT \cite{devlin19} and RoBERTa \cite{liu19}, as well as their variants BioBERT \cite{Lee19} and RoBERTa-Biomed \cite{gururangan20} fine-tuned on medical and scientific corpora. 

To highlight the impact of different fine-tuning strategies, we examine several configurations depending on the data and the masking strategy adopted. 
We experiment using the BioASQ QA training pairs during the fine-tuning stage and denote the models using them with \texttt{+BioASQ}.
%
When we fine-tune the models on the corpus \ml{consisting} 
of PubMed articles referred within the BioASQ and AI2's COVID-19 Open Research dataset, we compare two masking strategies denoted as \texttt{+STM} and  \texttt{+BEM}, where \texttt{+STM} indicates the standard masking strategy of the model at hand and \texttt{+BEM} is our proposed strategy. 
%
We additionally report the T5 \cite{raffel20} performance over 
CovidQA, which constitutes the current state-of-the-art~\cite{tang20}\footnote{We attach supplementary results in Appx.~\ref{sec:appendix} on SQuAD (Tab. \ref{tb:qa_results_full}) and the \textit{perplexity} of MLMs when fine-tuned on the medical collection with different masking strategies (Fig.~\ref{fig:perplexity_lms})}.


\noindent\textbf{Metrics}. To facilitate comparisons, we adopt the same evaluation scores used in \citet{tang20} to assess the models on the CovidQA dataset, i.e. mean reciprocal rank (MRR), precision at rank one (P@1), and recall at rank three (R@3); similarly, for the BioASQ dataset, we use the strict accuracy (SAcc), lenient accuracy (LAcc) and MRR, the BioASQ challenge's official metrics.

\section{Experimental Results and Discussion}
\label{sec:exp_results}
We report the results on the QA tasks in Table \ref{tb:qa_results}.

Among the unsupervised models, BERT achieves slightly better performance than RoBERTa on CovidQA, yet the situation is reversed on 
BioASQ (rows 1,5). The low precision of the two models (especially on the BioASQ dataset) confirms the difficulties in generalising to the biomedical domain. 
Specialised language models such as RoBERTa-Biomed and BioBERT show a significant improvement on the CovidQA dataset, but a rather limited one on BioASQ (rows 9,13), highlighting the importance of having larger medical corpora to assess the 
model's effectiveness. 
%
A general boost in performance is \ek{shared} across models fine-tuned on the QA tasks, with a large benefit from the BioASQ QA. \ek{The performance gains obtained by the specialised models (BioBERT and RoBERTa-Biomed) suggest} the importance of transferring not only the domain knowledge but also the ability to perform the QA task itself (rows 9,10; 13,14).

A further fine-tuning step before the training over the QA pairs has been proven beneficial for all of the models. The BEM masking strategy has significantly amplified the model's 
\ml{generalisability}, with an increased adaptation to the biomedical themes shown by the notable improvement in R@3 and MRR; with the R@3 outperforming the state-of-the-art results of T5 fine-tuned on MS-MARCO \cite{bajaj18} and proving the effectiveness of the BEM strategy.  


Table \ref{tab:qa_pairs_codvidqa} reports questions from the CovidQA related to three statistical 
\ml{indices} (i.e. Odds Ratio, Hazard Ratio and Relative Risk) to assess the risk of an event occurring in a group (e.g. infections or death). We notice that 
\ml{even though the indices are mentioned as abbreviations, }BERT fine-tuned with the STM is able to retrieve sentences with the exact answer for just one of three questions. 
\ml{By contrast}, BERT fine-tuned with the BEM strategy succeeds in retrieving at least one correct sentence for each question. 
This example suggests the importance of placing the emphasis on the entities, which might be overlooked by LMs during the training process despite being available.

\section{Related Work}

Our work is closely related to two lines of research: the design of masking strategies for LMs and the development of specialized models for the biomedical domain.

\noindent \textbf{Masking strategies.} Building on top of the BERT's masking strategy \cite{devlin19}, a wide variety of approaches has been proposed \cite{liu19, Yang19, jiang20}. 

A family of masking approaches aimed at leveraging entity and phrase occurrences in text. SpanBERT, \citet{Joshi20} proposed to mask and predict whole spans rather than standalone tokens and to make use of an auxiliary objective function. 
ERNIE \cite{zhang19} is instead developed to mask well-known named entities and phrases to improve the external knowledge encoded. Similarly, KnowBERT \cite{peters19} explicitly model entity spans and use an entity linker to an external knowledge base to form knowledge enhanced entity-span representations.
However, despite the analogies with the BEM approach, the above masking strategies were designed to generally improve the LM representations rather than adapting them to particular domains, requiring additional objective functions and memory.

\noindent \textbf{Biomedical LMs.} 
Particular attention has been devoted to the adaptation of LMs to the medical domain, with different corpora and tasks requiring tailored methodologies.
BioBERT \cite{Lee19} is a biomedical language model based on BERT-\textit{Base} with additional pre-training on biomedical documents from the PubMed and PMC collections using the same training settings adopted in BERT.
BioMed-RoBERTa \cite{gururangan20} is instead based on RoBERTa-\textit{Base} \cite{liu19} using a corpus of 2.27M articles from the  Semantic Scholar dataset \cite{ammar18}.
SciBERT \cite{beltagy19} follows the BERT's masking strategy to pre-train the model from scratch using a scientific corpus composed of papers from Semantic Scholar \cite{ammar18}. Out of the 1.14M papers used, more than $80\%$ belong to the biomedical domain.

\section{Conclusion}
We presented BEM, a biomedical entity-aware masking strategy to boost LM adaptation to low-resource biomedical QA. It uses an entity-drive\ml{n} masking strategy to fine-tune LMs and effectively lead them in learning entity-centric knowledge based on the pivotal entities characterizing the domain at hand. Experimental results have shown the benefits of such an approach on several metrics for biomedical QA tasks.

\section*{Acknowledgements}

This work is funded by the EPSRC (grant no. EP/T017112/1, EP/V048597/1). YH is supported by a Turing AI Fellowship funded by the UK Research and Innovation (UKRI) (grant no. EP/V020579/1).

\bibliography{anthology,eacl2021}
\bibliographystyle{acl_natbib}

\newpage
\onecolumn
\appendix

\section{Appendix}
\label{sec:appendix}

{\smallernormalAppendix 
We further examined whether the fine-tuning of the QA pairs affects not only the model adaptation to the QA task but it further helps realign the repression for the domain at hand.
The report scores point out that the vanilla LMs are the ones gaining the most when using in-domain QA pairs, such as BioASQ, compared to the SQuAD (rows 2,3; 9,10). The advantage tends to be reduced on already specialised LMs (rows 16,17; 23;24).
}

\vspace{20pt}

\setcounter{table}{0}
\renewcommand{\thetable}{A\arabic{table}}

\begin{table*}[!htb]
\centering
\scalebox{0.9}{
\begin{tabular*}{0.83\textwidth}{@{\extracolsep{\fill}}*{8}{>{\smallernormal}l}}
\toprule
\multirow{2}{*}{\textbf{\#}} & \multirow{2}{*}{\textbf{Model}}  & \multicolumn{3}{c}{\textbf{CovidQA}} & \multicolumn{3}{c}{\textbf{BioASQ 7b}} \\ \cline{3-5} \cline{6-8}  
                            &  & P@1 & R@3 & MRR           & SAcc & LAcc & MRR \\ \toprule
1 $\ \ $ & \textbf{BERT} 							& 0.081$^*$ & 0.117$^*$  & 0.159$^*$ & 0.012 & 0.032  & 0.027	\\
2 & $\ \ $ \texttt{+ SQuAD}                 & 0.110  	&  0.131  & 0.158	& 0.292  	& 0.343   & 0.318	\\
3 & $\ \ $ \texttt{+ BioASQ}                & 0.125		&  0.177  & 0.206  	& 0.226  	& 0.317   & 0.262  	\\
4 & $\ \ $ \texttt{+ STM + SQuAD}     		& 0.114  	&  0.146  & 0.173	& 0.305  	& 0.355   & 0.336	\\ 
5 & $\ \ $ \texttt{+ STM + BioASQ}         	& 0.132		&  0.195  & 0.218  	& 0.233  	& 0.325   & 0.265  	\\
6 & $\ \ $ \texttt{+ BEM + SQuAD}           & 0.126  	&  0.173  & 0.191	& 0.317 	& 0.371   & 0.349   \\ 
7 & $\ \ $ \texttt{+ BEM + BioASQ}          & 0.145     &  0.278  & 0.269  	& 0.241 	& 0.341   & 0.288  	\\ \hline

8 & \textbf{RoBERTa} 		   				& 0.068  	& 0.115	  & 0.122   & 0.023 	& 0.041  & 0.036   	\\
9 & $\ \ $ \texttt{+ SQuAD }                & 0.098 	& 0.134   & 0.160	& 0.353  	& 0.365  & 0.328   	\\
10 & $\ \ $ \texttt{+ BioASQ}                & 0.106		& 0.155   & 0.178	& 0.278  	& 0.324	 & 0.294	\\
11 & $\ \ $ \texttt{+ STM + SQuAD }          & 0.107 	& 0.148   & 0.175	& 0.361  	& 0.388  & 0.347   	\\
12 & $\ \ $ \texttt{+ STM + BioASQ}          & 0.112		& 0.167   & 0.194	& 0.282  	& 0.333	 & 0.300	\\
13 & $\ \ $ \texttt{+ BEM + SQuAD}           & 0.114 	& 0.162   & 0.185	& 0.368    	& 0.391  & 0.353   	\\   
14 & $\ \ $ \texttt{+ BEM + BioASQ}          & 0.125   	& 0.198   & 0.236 	& 0.323 	& 0.374  & 0.325   	\\ \hline

15 & \textbf{RoBERTa-Biomed}					& 0.104  	& 0.163 	& 0.192		& 0.028 	& 0.044  	& 0.037		\\
16 & $\ \ $ \texttt{+ SQuAD}                 & 0.111  	& 0.308 	& 0.288		& 0.376  	& 0.382   	& 0.358  	\\ 
17 & $\ \ $ \texttt{+ BioASQ}                & 0.128  	& 0.355  	& 0.315	   	& 0.415 	& 0.398		& 0.376		\\
18 & $\ \ $ \texttt{+ STM + SQuAD}           & 0.118  	& 0.314 	& 0.297		& 0.381  	& 0.390   	& 0.367  	\\ 
19 & $\ \ $ \texttt{+ STM + BioASQ}          & 0.136  	& 0.364  	& 0.321	   	& 0.423 	& 0.410		& 0.397		\\
20 & $\ \ $ \texttt{+ BEM + SQuAD}           & 0.121  	& 0.331 	& 0.323		& 0.385 	& 0.397   	& 0.378  	\\ 
21 & $\ \ $ \texttt{+ BEM + BioASQ}          & 0.143  	& 0.386  	& 0.347		& \textbf{0.435} 	& 0.443   	& 0.398  	\\ \hline

22 & \textbf{BioBERT}  		   				& 0.097$^*$ 	& 0.142$^*$ & 0.170$^*$ 	& 0.031 	& 0.046  & 0.039	\\
23 & $\ \ $ \texttt{+ SQuAD}                 & 0.161$^*$ 	& 0.403$^*$ & 0.336$^*$		& 0.381 	&  0.445  & 0.397  	\\
24 & $\ \ $ \texttt{+ BioASQ}                & 0.166 		& 0.419    	& 0.348	   		& 0.410$^\dag$  & 0.474$^\dag$  & 0.409$^\dag$	\\
25 & $\ \ $ \texttt{+ STM + SQuAD}           & 0.161 		& 0.411  	& 0.339	  		& 0.387   	&  0.447  & 0.401  \\ 
26 & $\ \ $ \texttt{+ STM + BioASQ}          & 0.172     	& 0.432  	& 0.385	  		& 0.418 	&  0.482  & 0.416  \\
27 & $\ \ $ \texttt{+ BEM + SQuAD}           & 0.168 		& 0.427  	& 0.354	  		& 0.391   	&  0.458  & 0.423  \\ 
28 & $\ \ $ \texttt{+ BEM + BioASQ} $\ \ \ $ & \textit{0.179}&\textbf{0.458} & \textit{0.391}     & 0.421 	&  \textbf{0.497}  & \textbf{0.434}  \\ \hline

29 & \textbf{T5 LM}  		   				   	\\	
30 & $\ \ $ \texttt{+ MS-MARCO} 			    & \textbf{0.282}$^*$ & 0.404$^*$ & \textbf{0.415}$^*$ 	& ---   	& ---     & ---		\\

\bottomrule
\end{tabular*}}
\caption{Performance of language models on the CovidQA and BioASQ 7b1 dataset. Values referenced with {\small*} comes from the \citet{tang20} work and with \dag $\,$ from \citet{Yoon20}.}
\label{tb:qa_results_full}
\end{table*}

\newpage
{\smallernormalAppendix 
In Figure \ref{fig:perplexity_lms}, we report the LM perplexity obtained when fine-tuning the model with the standard masking strategy versus the BEM strategy with different proportion of medical entities. Vanilla LMs experienced a huge gain with just a small fraction of entities, while already specialised LMs has a lower but still significant improvement. This could be expected as the specialised LMs has already encoded a large domain knowledge with representations that need to be realigned to the new ones.
}

\vspace{20pt}

\setcounter{figure}{0}
\renewcommand{\thefigure}{A\arabic{figure}}

\begin{figure*}[!htb]
\centering
\includegraphics[width=0.85\textwidth]{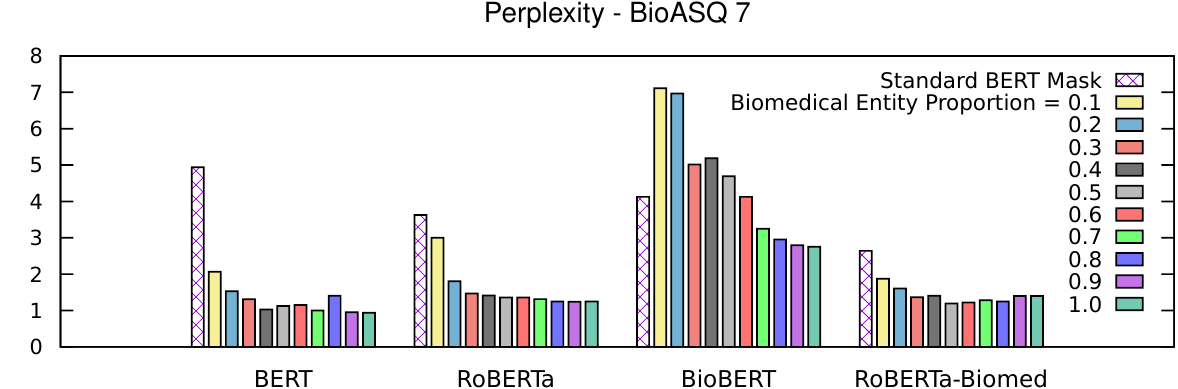}
\caption{Perplexity of MLMs using different masking strategies on the collection of medical articles.} 
\label{fig:perplexity_lms}
\end{figure*}

\end{document}